\documentclass{article}

\PassOptionsToPackage{numbers, compress}{natbib}
\usepackage[preprint]{neurips_data_2023}

\usepackage[utf8]{inputenc} 
\usepackage[T1]{fontenc}    
\usepackage{hyperref}       
\usepackage{url}            
\usepackage{booktabs}       
\usepackage{amsfonts}       
\usepackage{nicefrac}       
\usepackage{microtype}      
\usepackage{xcolor}         

\usepackage{graphicx}
\usepackage{wrapfig}
\usepackage{float}
\usepackage{subfig}
\usepackage{multirow}

\title{Infer Induced Sentiment of Comment Response to Video: A New Task, Dataset and Baseline}
\author{%
  Qi Jia$^{1}$\quad
  Baoyu Fan$^{2,1}$\footnotemark[1]\quad
  Cong Xu$^{1}$\quad
  Lu Liu$^{1}$\quad
  Liang Jin$^{1}$\quad
  Guoguang Du$^{1}$\quad
  Zhenhua Guo$^{1}$\quad 
  \And
  Yaqian Zhao$^{1}$\quad
  Xuanjing Huang$^{3}$\quad
  Rengang Li$^{4,1}$ \\
  $^1$Inspur Electronic Information Industry Co.,Ltd.\\ 
  $^2$Nankai University, 
  $^3$Fudan University, 
  $^4$Tsinghua University \\
  \small\texttt{jiaqi01@ieisystem.com, fanbaoyu@foxmail.com}, \\
  \small\texttt{\{xucong, liulu06, jinliang, duguoguang, guozhenhua\}@ieisystem.com},\\
  \small\texttt{zhaoyaqian@ieee.org, xjhuang@fudan.edu.cn, lirengangx@gmail.com}\\
}

\begin{document}

\maketitle

\begin{abstract}
Existing video multi-modal sentiment analysis mainly focuses on the sentiment expression of people within the video, yet often neglects the induced sentiment of viewers while watching the videos. 
Induced sentiment of viewers is essential for inferring the public response to videos, has broad application in analyzing public societal sentiment, effectiveness of advertising and other areas. 
The micro videos and the related comments provide a rich application scenario for viewers' induced sentiment analysis.
In light of this, we introduces a novel research task, \textbf{M}ulti-modal \textbf{S}entiment \textbf{A}nalysis for \textbf{C}omment \textbf{R}esponse of \textbf{V}ideo \textbf{I}nduced(\textbf{MSA-CRVI}), aims to inferring opinions and emotions according to the comments response to micro video.  
Meanwhile, we manually annotate a dataset named \textbf{C}omment \textbf{S}entiment  toward to \textbf{M}icro \textbf{V}ideo (\textbf{CSMV}) to support this research. It is the largest video multi-modal sentiment dataset in terms of scale and video duration to our knowledge, containing $107,267$ comments and $8,210$ micro videos with a video duration of 68.83 hours.
To infer the induced sentiment of comment should leverage the video content, so we propose the \textbf{V}ideo \textbf{C}ontent-aware \textbf{C}omment \textbf{S}entiment \textbf{A}nalysis (\textbf{VC-CSA}) method as baseline to address the challenges inherent in this new task. 
Extensive experiments demonstrate that our method is showing significant improvements over other established baselines.
We make the dataset and source code publicly available at \url{https://github.com/AnonymousUserabc/MVI-MSA_DateAndSource.git}.
\end{abstract}

\section{Introduction}
\label{sec:intro}

\begin{figure}
  \centering
  \includegraphics[width=0.85\textwidth]{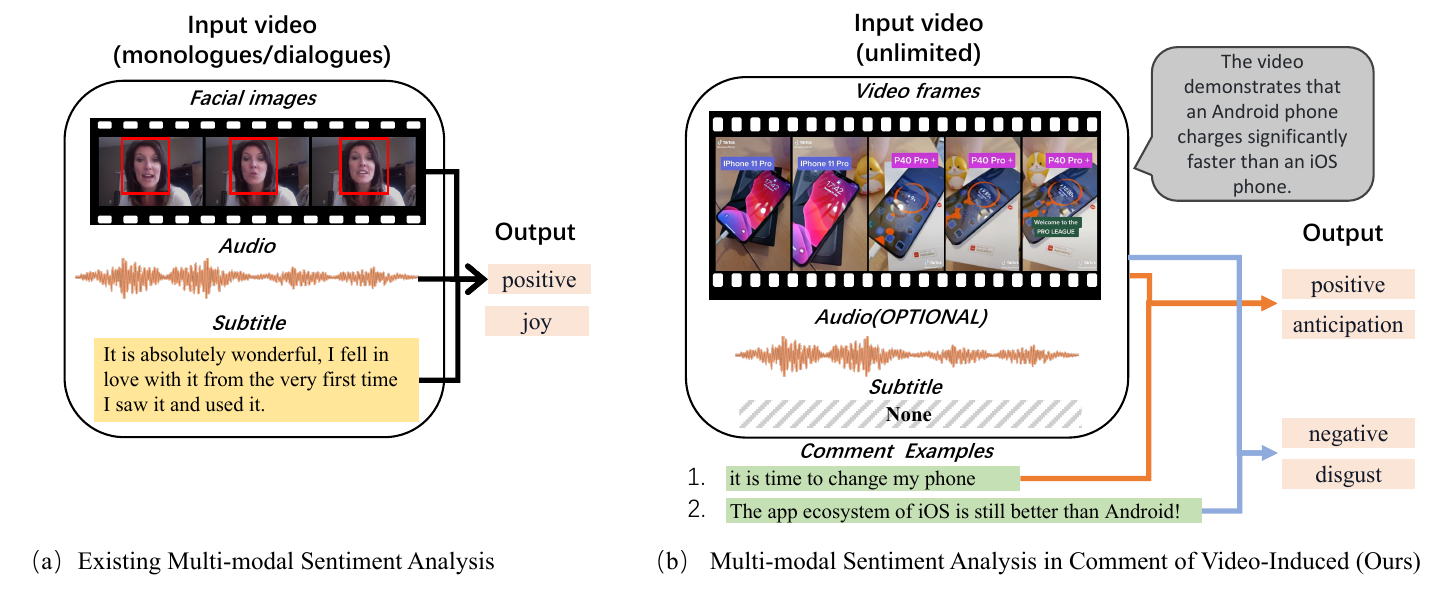}
  \caption{Figure (a) describes the setting of traditional multi-modal sentiment analysis, which aims to determine the speaker's sentiment based on the given multi-modal information. 
  Figure (b) illustrates the example of our proposed task.
  Two comments are highlighted in the figure and hold different induced sentiments toward the related video. 
  For easy comprehension, a description of the video content is presented in a gray box.
  This description does not serve as input.}
  \label{fig1-instance}
\end{figure}

Video multi-modal sentiment analysis, a captivating and challenging research field, has exhibited rapid advancements with a variety of benchmarks proposed in recent years~\cite{PrezRosas2013UtteranceLevelMS, 8489099, liu2022make, zadeh2018multimodal, busso2008iemocap, poria-etal-2019-meld}. 
These benchmarks aim to understand the opinions or emotions of speakers in monologues or dialogues, as depicted in Fig.~\ref{fig1-instance}a. 
They consider the combined input from visual, audio, and subtitle text at the utterance level, which maintain the same semantic(\textit{"It is absolutely wonderful, I fell in love with it from the very first time I saw it and used it."}).
Current methods infer the speaker's opinions or emotions by examining elements such as sequential images (e.g., facial expressions, smiles, gazes), audio cues (e.g., tones, pauses, pitch), and transcribed text from spoken words~\cite{GANDHI2023424}. 

Nevertheless, current research has primarily centered on the sentiments of the people in the video, paying less attention to viewers' induced sentiment while watching the video.
People create and upload micro videos, and viewers contribute comments as responses to the micro video ~\cite{10.1145/3479574}. 
These comments often reveal sentiments which are induced by the video.
Analyzing the viewers' induced sentiments of the video based on comments is significant for developing comprehensive applications such as the analysis of public societal sentiment, evaluating advertising effectiveness~\cite{Xu2021TikTokAP, SOUTHWICK2021234, LOVETT2021126, qiyang2019learning, 10.1145/3394231.3397916, aup:/content/journals/10.5117/CCR2022.2.004.GUIN, doi:10.1080/1057610X.2020.1780027}.
We consider the video-induced sentiment analysis as a new paradigm of multi-modal sentiment analysis.
In contrast to the existing multi-modal sentiment analysis, a single micro video may yield multitude comments, and each comment may express different sentiments about the video content.
As the example in Fig.~\ref{fig1-instance}b, the video shows that an android phone charge faster than an iOS phone, aim to illustrate the advantage of android phone. 
Comment 1 expresses a willingness to switch phone, thus agree with the related video. 
But the comment 2 offers disagreement by praising the better app ecosystem of iOS.
Obviously, just relying on text to infer comments' opinions and emotions toward the video is often inaccurate as both comments exhibit preferences for different smartphones. 
For precise sentiment inference from comments, it is essential to integrate the video content.
Current approaches tend to treat comment sentiment analysis to be a simple NLP task and neglect the semantic connection between videos and comments~\cite{YASMINA2016292,Uryupina2014SenTubeAC,Pokharel2021ClassifyingYC,oro32459,Alhujaili2021SentimentAF,Muhammad2019SentimentAO}. 

Considering this, we introduce a new task termed \textbf{M}ulti-modal \textbf{S}entiment \textbf{A}nalysis for \textbf{C}omment \textbf{R}esponse of \textbf{V}ideo \textbf{I}nduced(\textbf{MSA-CRVI}).
This task focuses on understanding the induced sentiment of the video, as conveyed through viewers' comments.
MSA-CRVI incorporates both the textual comment and the associated video as inputs. 
Unlike existing video multi-modal sentiment analysis, MSA-CRVI task presents unique challenges within this innovative paradigm.
Firstly, it is challenging to ground the associated video content with each comment.
A single video yields a multitude comments that emphasize diverse aspects, necessitating grounding the relevant video contents for each comment's sentiment analysis.
Since comments are responses to the video rather than mere textual descriptions, it becomes challenging to directly grounding the video content with the comment.
Secondly, it is challenging to model the correlation between comments and their corresponding micro videos due to its temporal complexity.
Comments could focus on different temporal-granularity content within the video.
Meanwhile, comment may be grounding multiple segment across various video timeline. 
This implies the necessity for carefully encoding video temporal features and precisely processing the grounding video information.

We have developed a dataset to support the MSA-CRVI task, called \textbf{C}omment \textbf{S}entiment toward \textbf{M}icro \textbf{V}ideo (\textbf{CSMV}), collected from TikTok, a popular micro video social media platform. CSMV comprises micro videos and associated comments, each of which is annotated for opinions and emotions. 
Furthermore, we propose a strong baseline method, named \textbf{V}ideo \textbf{C}ontent-aware \textbf{C}omment \textbf{S}entiment \textbf{A}nalysis (\textbf{VC-CSA}) to address these challenges by designing three key modules: Multi-scale Temporal Representation, Consensus Semantic Learning and Golden Feature Grounding.
Comprehensive experiments have validated that our method significantly outperforms established baselines.
The data and source code are released at \url{https://github.com/AnonymousUserabc/MVI-MSA_DateAndSource.git}.

Our main contributions including 
(1) We introduce the \textbf{MSA-CRVI} task with a novel setting in multi-modal sentiment analysis. This task involves inferring the induced sentiment according to the comments toward micro-video.
(2) To support this task, we have created a dataset named \textbf{CSMV}, comprising manually annotated opinions/emotions on comments and related videos. To our knowledge, CSMV is the largest dataset of its kind in terms of scale and video duration.
(3) As an initial exploration of the task, we present the \textbf{VC-CSA} method, which focus on understanding the correlation between comments and micro-videos to infer the opinions and emotions induced by the videos.
(4) The extensive experiments demonstrated that VC-CSA outperforms other state-of-the-art multi-modal sentiment analysis methods on the CSMV dataset. We also highlight the critical role of video in the MSA-CRVI task.

\section{Related work}
\label{sec:relatedwork}
\textbf{Multi-modal sentiment analysis} has gained substantial attention and driven the rapid expansion of multi-modal applications.
Over the years, a variety of multi-modal datasets have emerged, typically categorized based on the presentation of videos.
One category comprises datasets featuring monologue-style videos, such as MOUD~\cite{PrezRosas2013UtteranceLevelMS}, OMG-Emotion~\cite{8489099}, CH-SIMS~\cite{yu-etal-2020-ch,liu2022make}, CMU-MOSEI~\cite{zadeh2018multimodal} and so on. 
Another category encompasses dialogue-style video datasets like IEMOCAP~\cite{busso2008iemocap}, MELD~\cite{poria-etal-2019-meld}, often derived from movies and TV shows. 
Existing multi-modal sentiment analysis datasets impose stringent presentation constraints, but micro-videos is much more diverse in content and format than monologue and dialogue.

Many approaches have been proposed for the multi-modal sentiment analysis.
Md Shad Akhtar et al.~\cite{akhtar2019multi} 
introduced a context-level inter-modal attention framework aiming to infer the sentiment expressed by the speaker utterance.
Delbrouck et al.~\cite{delbrouck2020transformer} proposed a transformer-based joint-encoding method employing cross-modal attention mechanisms to capture inter-modality interactions. 
Their experimental verification revealed the limited role of the visual modality in reasoning within existing multi-modal sentiment analysis.
Subsequent studies, including MMIM~\cite{han2021improving}, Self-MM~\cite{yu2021learning}, and MISA~\cite{hazarika2020misa}, further support this perspective.
Obviously, researchers focus on fusing signals from varied modalities to extract complementary information sharing the same semantic meaning.
However, these relevant approaches are confined by the current setting which is same as the constraints of the benchmark.
Different from the prior research, VI-MSA task in this paper provides a more complex analysis situation between the video and the textual comment. 
Additionally, the video themes exhibit greater diversity and a freer style.

\textbf{Induced emotion analysis}, distinct from perceiving emotion conveyed by content creators, pertains to analyzing emotional reactions induced from content consumers~\cite{kallinen2006emotion, tian2017recognizing}. 
Presently, there is a growing interest in comprehending the patterns of emotion induced by video~\cite{baveye2013large}, since it has a wide range of applications in various perspectives~\cite{teixeira2012emotion,sukhwal2022determining,peng2015mixed,alqahtani2022predicting}.
Currently, researchers mainly investigate induced emotion of viewer responses to movies (e.g., DEAP~\cite{5871728}, COGNIMUSE~\cite{Zlatintsi2017COGNIMUSEAM}, LIRIS-ACCEDE~\cite{baveye2013large}). 
They capture viewers' physiological features, like EEG and facial videos, to facilitate the induced emotion analysis~\cite{5871728,5975141}.
Typically, these datasets offer continuous numerical labels for arousal and valence.
Due to these characteristics, theses dataset is expensive in construction cost and severely restricted for application. 
In comparison, micro videos are largely created in a freestyle, and the related comments are often easy to obtain and directly reflect induced sentiment.

Several attempts have been made to infer the induced emotions of videos.
Benini et al.~\cite{benini2011connotative} argued that similar connotations in movie scenes could evoke identical emotional responses. 
They proposed a method to develop a construct for affective description for movies based on their connotative properties.
Tian et al.~\cite{tian2017recognizing} emphasized the difference between perceived and induced emotions in the viewer. 
They employed an LSTM-based model to recognize induced emotions from the viewers' physiological features, showcasing the effect of integrating multiple modalities, including external information like affective cues in movies.
Muszyński et al.~\cite{muszynski2019recognizing} further investigated the correlation between dialogue and aesthetic features in inducing emotions in movies. They introduced an innovative multi-modal model for predicting induced emotions.
Liu et al.~\cite{liu2017real} employed EEG signals to real-time infer induced emotions in audiences while watching movies. Their study centered on the widely used LIRIS-ACCEDE database~\cite{baveye2013large} in recent research on induced emotions in movies. 
These studies aimed to advance the movie art research and aid filmmakers in creating emotionally engaging content, where features beyond video are employed to infer induced emotions. 
These methodologies mainly rely on the viewers' physiological features, which significantly differ from the video-induced multi-modal sentiment analysis task that utilizes textual comments and videos.

\section{Dataset}
\label{dataset}

\subsection{Data collection}

\textbf{TikTok} is one of the most popular micro video social media platforms. 
Users spontaneously create micro videos and contribute related comments as responses on TikTok.
These videos encompass diverse topics (e.g., sports, politics, technology), reflecting human experiences, thereby provide a substantial amount of valuable data for our proposed task MSA-CRVI.
The metadata of micro videos on TikTok includes hashtags denoting video topics and the number of likes on comments, facilitating raw data processing~\cite{10.1145/3479574}.

We employ hashtags to collect raw data from Tiktok.
Hashtags are formed spontaneously by users creating micro videos, reflect current trends on social media platforms.
To enhance data diversity, we set many hashtags with different topics and no restrictions on micro video representation format. 
A set of hashtags encompassing diverse topics like policy, business, sports, and technology is manually selected.
For ensuring the quality of micro videos and comments, micro videos with less than $1,000$ comments are excluded.
Then, we sort comments for each micro video based on the number of likes and select the top 20 English comments for annotation. 
Furthermore, a series of pre-processing steps is undertaken to prevent personal information leakage. 
Initially, we delete the metadata about the creators of micro videos and comments. 
Subsequently, any personal information within textual comments (e.g., usernames, emails, phone numbers) is removed. 
Lastly, instead of the raw video data, micro video features generated via the pre-trained visual model I3D~\cite{carreira2017quo} are published. 
The same features will serve to evaluate our proposed method. 

\subsection{Data annotation}

\begin{table}[ht]
\centering
\caption{The annotation guidelines for labeling comments on micro videos.}
\scalebox{0.8}[0.8]{
\begin{tabular}{c|l|l}
\hline
\multicolumn{1}{l|}{Task} &
  Label &
  Description \\ \hline
\multirow{3}{*} {Opinion} &
  positive &
  \begin{tabular}[c]{@{}l@{}}Hold a positive attitude towards the content of the video, agree with the information \\
  presented in the video, consider the video to be accurate, and experience a sense of \\comfort induced by the video.\end{tabular} \\
 &
 negative &
  \begin{tabular}[c]{@{}l@{}}Hold a negative attitude towards the content of the video, disagree with the information\\ 
  presented in the video, consider there to be errors in the video, and feel uncomfortable \\because of the video.\end{tabular} \\
 &
  neutral &
  \begin{tabular}[c]{@{}l@{}}Hold no clear bias towards the content of the video; provide objective statements without\\
  any particular leaning; make comments that are associations triggered by the video rather \\than expressing a specific
  attitude; make comments that are not directly related to \\the content of the video.\end{tabular} \\ \hline
\multirow{8}{*} {Emotion} &
  fear &
  \begin{tabular}[c]{@{}l@{}}Fear, terror, apprehension evoked by the video, including reactions of being startled by\\ watching
  the video, etc.\end{tabular} \\
 &
  disgust &
  Disgust, dislike, boredom for video content, uninterested in video. \\
 &
  anger &
  Rage, anger, annoyance cause by the video. \\
 &
  sadness &
  Feel sadness, grief within the video. Catch pensiveness in video. \\
 &
  joy &
  \begin{tabular}[c]{@{}l@{}}Feel happy, joyful, or serenity in heart because of video, including teasing and laughing \\at the content of the video\end{tabular} \\
 &
  trust &
  \begin{tabular}[c]{@{}l@{}}Trust, or feel admiration, or express a convinced attitude towards the content of the video.\end{tabular} \\
 &
  anticipation &
  \begin{tabular}[c]{@{}l@{}}Looking forward to, sparking curiosity about, or expressing anticipation cause of the video.\end{tabular} \\
 &
  surprise &
  \begin{tabular}[c]{@{}l@{}}The content of the video is surprising, amazed, or shocked more than expected.\end{tabular} \\ \hline
\end{tabular}
}
\label{labelDefinition}
\end{table}

We employed 30 human annotators to manually label comments, defining two distinct types of labels: opinion and emotion.
The opinion label indicates the user's attitude towards the micro video in comment.
This can encompass agreement with or expression of feelings towards the video, ranging from positive, negative, to neutral.
Specifically, the neutral label signifies an absence of clear opinion or views unrelated to the video.
The emotion label illustrates the emotional reaction in a comment evoked by the micro video.
We employ the Plutchik wheel~\cite{PLUTCHIK19803} to define eight categories: joy, disgust, surprise, sadness, trust, fear, anger, and anticipation. 
These categories encompass a wide range of emotional directions, each illuminated into three levels from mild to intense, effectively capturing human emotional expressions.

We devised a data annotation workflow to ensure annotator quality and reduce individual subjective biases in the annotations.
Detailed guidelines and processes for annotating our CSMV dataset are outlined below.

\textbf{Annotation guidelines.} Initially, we establish comprehensive annotation guidelines that precisely define the criteria for data labeling (referenced in Tab.~\ref{labelDefinition}).
These guidelines provide explicit instructions on for identifying and annotating various elements within the dataset.
    
\textbf{Pre-annotation phase.} Prior to formal annotation, we create a small dataset with ground truth to select annotators. We request them to undergo three rounds of annotation. After each round, we review the results across all annotators. On the final round, annotators with labeling accuracy exceeding $90\%$ are chosen for the formal annotation.
    
\textbf{Formal annotation phase.} The raw data comprises both of the comments and the related micro videos. We allocate the raw data to each annotator on a hashtag-level. 
Annotators are tasked with simultaneously labeling opinion and emotion for each comment.
Besides, we emphasize the comment which difficult to understand should be skipped. 
Throughout this phase, we maintain communication and offer support to annotators. 
Concurrently, regular meetings and feedback sessions facilitate continuous improvement and uphold high-quality annotations.
    
\textbf{Cross-validation.} 
Assessing opinions and emotions often entails subjective judgment.
To reduce personal bias, we implement a 3-fold cross-validation among annotators.
We randomly sample 20$\%$ of the labeled data from each hashtag and exchange them among two other annotators for correction. 
Subsequently, consistency is calculated based on the validation outcomes. 
If the rate of inconsistent labels surpasses 10$\%$, annotators are requested to re-label the comment data of the hashtag.
This cross-validation process safeguards labeling consistency.

Finally, we construct \textbf{CSMV} dataset comprising $107,267$ comments and $8,210$ micro videos collected from 35 hashtags, totaling a video duration of 68.83 hours.

\subsection{Comparison of dataset statistics}

\begin{table}[t]
\centering
\caption{
The statistical information of datasets include video induced emotion, multi-modal sentiment analysis, and the proposed dataset CSMA, from top to bottom.
}
\scalebox{0.80}{
\begin{tabular}{lllll}
\hline
Dataset      & Scale       & {Video Duration} & {Video Representation}     \\ \hline
DEAP~\cite{5871728}         & $120$       & 2 hours           & music          \\
COGNIMUSE~\cite{Zlatintsi2017COGNIMUSEAM}    & $50$          & 3.5 hours         & movie        \\
LIRIS-ACCEDE~\cite{baveye2013large} & $9,800$      & 27 hours          & movie          \\ \hline
MOUD~\cite{PrezRosas2013UtteranceLevelMS}         & $400$        & 1 hour            & monologue      \\
OMG-Emotion~\cite{8489099}  & $2,400$      & 1 hour            & monologue      \\
CH-SIMS2.0~\cite{liu2022make}   & $4,402$      & 4.43 hours        & monologue      \\
CMU-MOSEI~\cite{zadeh2018multimodal}    & $23,453$     & 65.9 hours        & monologue      \\
IEMOCAP~\cite{busso2008iemocap}      & $7,433$      & 12 hours          & dialogue       \\
MELD~\cite{poria-etal-2019-meld}         & $13,000$     & 13 hours          & dialogue       \\ \hline
\textbf{CSMV}         & \textbf{107,267}    & \textbf{68.83 hours}       & \textbf{unlimited}      \\ \hline
\end{tabular}
}
\label{datasetcomp}
\end{table}


Tab.~\ref{datasetcomp} presents a comparison between CSMV and current multi-modal sentiment analysis datasets. It provides details about scale, video duration, and video content on each dataset. 
In terms of scale, CSMV stands out with a substantial sample count of 107,267. 
Meanwhile, the video duration of CSMV is 68.83 hours, offering notably extensive video content.
This indicates that CSMV provide the a relatively large scale both in scale and video duration.
Furthermore, a key distinction lies in is the video representation in CSMV is unlimited. 
Comparatively, the existing video induced emotion focus on the movie, and the existing multi-modal dataset focus on the sentiment of speaker in the video.
These datasets have limitations in conveying visual information and expression.
Conversely, our proposed CSMV comprises a broader and more diverse range of video representation, potentially introducing additional complexities and challenges in sentiment analysis.
More statistics pertaining to our CSMV dataset are available in the supplementary materials.


\subsection{Ethics}

Concerning personal privacy, the CSMV dataset would not publish the original videos. Instead, it publishes only the visual features extracted from micro videos using the pre-trained I3D model~\cite{carreira2017quo}.
Additionally, the comments solely preserve the text, removing all user-related information. 
Both the code and data are publicly accessible under the CC BY-NC-SA 4.0 license, intended for academic and non-commercial use.

\section{Method} \label{method}

\begin{figure}[t]
  \centering
  \includegraphics[width=0.9\linewidth]{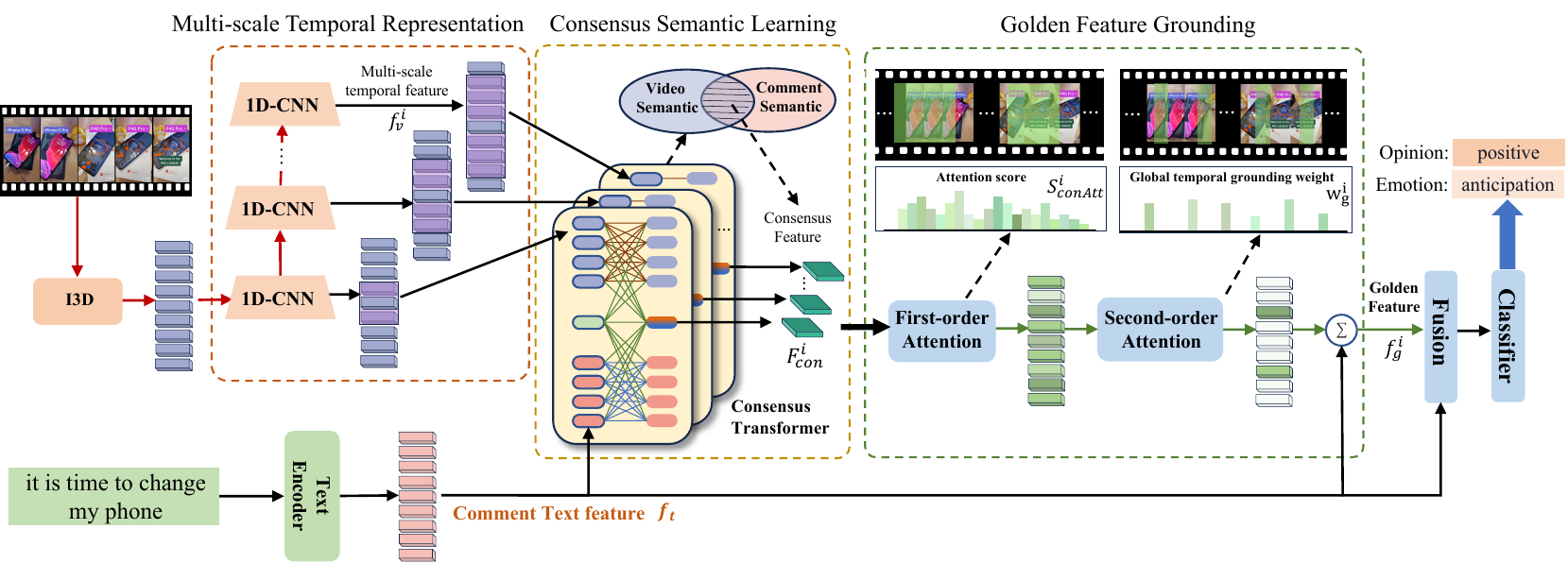} 
  \caption{The architecture of \textbf{V}ideo \textbf{C}ontent-aware \textbf{C}omment \textbf{S}entiment \textbf{A}nalysis (\textbf{VC-CSA}).
  We mainly design Multi-scale Temporal Representation, Consensus Semantic Learning and Golden Feature Grounding modules to address the new challenges of the proposed task.} 
  \label{method-overview} 
\end{figure}

To infer the comment’s induced sentiment toward to related micro video, we propose a novel method called \textbf{V}ideo \textbf{C}ontent-aware \textbf{C}omment \textbf{S}entiment \textbf{A}nalysis (\textbf{VC-CSA}).
It takes a comment and the related micro video  as input to infer the opinion and emotion toward to the video which expressed through the comment. Fig.~\ref{method-overview} is the architecture of the framework. 
The visual encoder is a video pre-trained model I3D~\cite{carreira2017quo}, which encodes the micro video into a set of vector representations as original temporal visual features input. 
The comment text is encoded by a RoBERTa~\cite{liu2019roberta} language pre-trained model to extract text features from the comment. 
Our proposed method consists of three principal modules: Multi-scale Temporal Representation, Consensus Semantic Learning, and Golden Feature Grounding. 
We integrate the multi-scale video golden feature with the textual comment with a fusion module and utilize a $Softmax$ classifier to infer opinions and emotions. 
For training, we apply cross-entropy loss to each classification head and aggregate these losses for optimization.

\subsection{Multi-scale Temporal Representation}
Each video has the potential to provoke a multitude of comments from viewers. 
These comments may address specific segments or the entirety of the video story. 
For instance, in a video where a dog chases a cat and ultimately collides with a door, one comment, '\textit{That hurts},' refers to the end of video. 
In contrast, another comment, '\textit{Dogs do not like cats},' reflect the general theme of the video. 
Consequently, it is crucial to encode the semantic features of the video across various  temporal scales to facilitate the correlation between the video and comments spanning different different time ranges.

The \textbf{Multi-scale Temporal Representation} module is design to capture the visual features from the video in various temporal scales. 
This is accomplished by stacking multiple 1D Convolutional Neural Networks (CNNs) and employing the $ReLU$ activation function between layers.
Each 1D-CNN operates with a kernel size of three and a stride of one, traversing the temporal dimension of the video's visual features input.
As the number of layers increases, the network progressively expand broader temporal contexts within the video's visual features.
Consequently, we obtain hierarchy of multi-scale temporal representations, denoted as $\{{f_v^i} \in \mathbb{R}^{v_l\times d_v}\}$ of the video visual information, capturing a spectrum from finer to coarser granularities, where $v_l$ represents the length of the video,  $d_v$ denotes the dimension of the visual representations, and $i$ indexes the layers.. 
This approach is instrumental in analyzing the video's contextual representation difference over various time spans.

\subsection{Consensus Semantic Learning}
The comments, being responses to a video, do not describe the video content directly, creating a semantic gap between the video and the comments.
This distinction hinder directly grounding relevant video content  based on the comment text. 
To address this challenge, we introduce \textbf{Consensus Semantic Learning}, a module designed to deeply model semantic correlation between the video and the comments, facilitating more effective video content grounding. 

The foundation for effectively grounding video content lies in construct a well-defined query to bridge the semantic gap between the comments and videos. 
Therefore, we introduce a video-comment consensus transformer to capture the shared semantic occurrences between comment and video. 
As illustrated in Fig.~\ref{method-overview}, this transformer is a special structure derived from transformer encoder block. 
It processes video feature $f_v^i$, text feature $f_t$, and several trainable Consensus Tokens as input. 
Consensus token representations $f_{con} \in \mathbb{R}^{v_l\times d^T}$, are randomly initialized and become trainable in training phase, $d^T$ denotes the dimension of the consensus transformer. 
With in the construction, the attention connection between the video and text features are masked, allowing for information exchange solely through the Consensus Token. 
Consequently, the consensus tokens serve as mediums, sharing the semantic between the video and the comment. 
We take the consensus tokens representation output $F_{con}^{i}$ in Eq.~\ref{ConTrans} as a consensus feature to reduce the semantic gap like a bridge.
\begin{equation}
        F_{con}^{i} = ConsensusTansformer_{cons}([f_v^i;f_{con};f_t]) \label{ConTrans}
\end{equation}
In Eq.~\ref{ConTrans}, $f_t \in \mathbb{R}^{l_t \times d_t}$ denotes the comment text feature, where $l_t$ is the length of text, $d_t$ is the dimension. 
Importantly, the consensus transformer block is parameter-independent for each temporal scale of $\{{f_v^i}\}$. 
This methodology is applied across all multi-scale temporal features.

\subsection{Golden Feature Grounding}
To accurately interpret the sentiment of comment related to a video, it is important to ground the video content referenced by the comment. 
Given the temporal nature of video, continuous frames often exhibit high similarity, leading to redundant information during the grounding phase. 
To control this, We design \textbf{Golden Feature Grounding} module, which comprises a two-steps approach to compute grounding weight.
In the first-order grounding, we employ a multi-head attention mechanism. This mechanism  utilize the consensus token representation $F_{con}^{i}$ as the query, the video visual feature $f_v^i$ as the key and the value.
The process of calculation is illustrated as Eq.~\ref{MHA1}:
\begin{equation}
        {S_{conAtt}^{i}} = MultiHeadAttention(F_{con},f_v^i,f_v^i) \label{MHA1}
\end{equation}
Here, the attention score $S_{conAtt}^{i} \in \mathbb{R}^{{head} \times v_l}$ reflects the comment attention across the video temporal, with $head$ representing the number of attention heads.

Viewers' attention to video content varies over time, often focusing on multiple segments simultaneously. 
This attention score $S_{conAtt}^{i}$ may exhibit smoothness due to the similarity for adjacent temporal segments, resulting from the the temporal nature of video. 
Consequently, shorter segments could be overshadowed by longer ones if the attention score $S_{conAtt}^{i}$ is use directly to obtain video information relevant to the comment.
Address this, we design a second-order grounding to filter out the redundancies and obtain the golden feature that represents the essence of video relevant to the comment.
We take the multi-head attention score $S_{conAtt}^{i}$ as an indicator of temporal attention trends across distinct vector space. 
This score is input to a memory module to analyze the trend of attention scores in the temporal direction. 
The memory module is same as the the cell state of LSTM. 
Then, the representation is processed through a $ReLU$ function to obtain the global temporal grounding weight ${W_g^i} \in \mathbb{R}^{v_l}$.
\begin{equation}
        {W_g^i} = ReLU(cellState(LSTM(S_{conAtt}^{i})) \label{lstmrelu}
\end{equation}
This grounding weight is then multiply to the video features ${f_{v}^{i}}$ to produce the features ${f_{g}^i} \in \mathbb{R}^{d_v}$ along the temporal axis, which we regard as the golden features related to the comment. 
The calculation process is shown as Eq.~\ref{glodenclue}:
\begin{equation}
        f_{g}^i = \sum{W_g^i}{f_{v}^{i}} \label{glodenclue}
\end{equation}

\subsection{Fusion and Classifier}
After the steps outlined above, we introduce a fusion module designed to integrate video features across various temporal scales into the comment feature, thereby enriching the interaction between comment text and video data. 
To facilitate this, we employ a multi-view attention mechanism, wherein the comment text token feature denoted as $f_t^j$ as the query, and the video golden features in multi-scales, represent by $\{f_{g}^i\}$ as key and value. 
This approach specifically targets capturing the interactions at the token level between the comment and the video, where $j$ corresponds to the index of the token within the comment.
\begin{equation}
        {AttnScale}_j^i = Attention(f_t^j, \{f_{g}^i\}) \label{attnScale}
\end{equation}
\begin{equation}
        F_{g}^j = \sum_i{{AttnScale_j^i}{f_{g}^i}} \label{Fusion}
\end{equation}
Subsequently, we concatenate the features $\{F_{g}^j\} \in \mathbb{R}^{l_t \times d_v} $ and text features $f_t$ in token-level along the feature dimension to generated the video context-aware comment semantic feature $ f_s \in \mathbb{R}^{l_t \times (d_v+d_t)}$. 
This feature $f_s$ is processed through a layer of multi-head self-attention and a pooling mechanism to get the final context-aware comment semantic feature representation $F_s$ for sentiment analysis.
\begin{equation}
F_s=MaxPool(MultiHeadSelfAttention([\{F_{g}^j\};f_t])\label{Fs}
\end{equation}
This fusion strategy aims to incorporate both video and textual information into a unified representation. 
We utilize two $softmax$ functions on $F_s$ to calculate the possibility of opinion and emotion which the comment response to video. 

\section{Experiments} 
\label{sec:experiment}
We select representative sentiment analysis methods for comparison. 
Notably, our selection included methods that primarily utilize textual input, such as BERT ~\cite{devlin2018bert} and RoBERTa ~\cite{liu2019roberta}. 
We exclusively trained these models on the comment text from the CSMV dataset, facilitating an evaluation of the micro videos' impact on the MSA-CRVI task.
Furthermore, we select several typical traditional multi-modal sentiment analysis methods: TBJE ~\cite{delbrouck2020transformer}, SELF-MM~\cite{yu2021learning}, MISA~\cite{hazarika2020misa}, MMIM~\cite{han2021improving} and CubeMLP~\cite{10.1145/3503161.3548025}. 
Our method and comparative methods are implemented on the PyTorch platform~\cite{NEURIPS2019_9015} and trained on 4 Nvidia Tesla V100 GPUs.
Video visual features were integrated using the I3D model~\cite{carreira2017quo}.
For the implementation of our proposed model, we set the hidden dimensions $d_v$, $d^T$ and $d_t$ to 768.
To ensure equitable comparisons, we align the training settings (e.g., loss function, batch size, learning rate strategy, etc) with all methods.
To evaluate the performance of the models, we randomly split our dataset into training, development (dev), and testing sets using a ratio of 7:1:2. 
The dev set serves as the basis for selecting the most effective model for each method based on performance outcomes. 
We follow prevailing evaluation protocols to use F1-score as the primary metrics to measure the performance. 
Additionally, we calculate mean values from 5 random seeds for each performance metric.

\textbf{Comparison Analysis.}
\begin{table}[t]
\centering
\caption{The experiment results of the comparison.}
\scalebox{0.8}[0.8]{
    \begin{tabular}{c|cccccccc}
    \hline
    \multirow{3}{*} {Models} & \multicolumn{4}{c|}{Opinion}            & \multicolumn{4}{c}{Emotion}              \\ \cline{2-9}
                            & \multicolumn{1}{c|}{Micro}    & \multicolumn{3}{c|}{Macro}     & \multicolumn{1}{c|}{Micro}    & \multicolumn{3}{c}{Macro}     \\ \cline{2-9}
                            & \multicolumn{1}{c|}{F1-score} & F1-score & Recall & \multicolumn{1}{c|}{Precision} & \multicolumn{1}{c|}{F1-score} & F1-score & Recall & \multicolumn{1}{c}{Precision} \\
    \hline
    BERT \cite{devlin2018bert}(only text)	&\multicolumn{1}{c|}{56.42}&	48.52&	48.14&	\multicolumn{1}{c|}{49.31}&	\multicolumn{1}{c|}{43.34}&	33.64&	32.98&	34.59\\
    RoBERTa \cite{liu2019roberta}(only text)	&\multicolumn{1}{c|}{56.95}&	49.29&	48.87&	\multicolumn{1}{c|}{49.98}&	\multicolumn{1}{c|}{47.27}&	37.56&	36.85&	38.77\\ 
    \hline
    TBJE \cite{delbrouck2020transformer}	&\multicolumn{1}{c|}{65.81}&	59.80&	59.20&	\multicolumn{1}{c|}{60.94}&	\multicolumn{1}{c|}{55.67}&	48.14&	48.71&	46.61\\
    SELF-MM \cite{yu2021learning}	&\multicolumn{1}{c|}{65.77}&	58.56&	57.30&	\multicolumn{1}{c|}{61.20}&	\multicolumn{1}{c|}{53.92}&	46.44&	44.64&	49.87\\
    MISA \cite{hazarika2020misa}	&\multicolumn{1}{c|}{72.41}&	66.54&	65.40&	\multicolumn{1}{c|}{68.69}&	\multicolumn{1}{c|}{57.42}&	49.71&	48.07&	52.77\\
    MMIM \cite{han2021improving}	&\multicolumn{1}{c|}{65.40}&	58.39&	59.96&	\multicolumn{1}{c|}{57.65}&	\multicolumn{1}{c|}{52.35}&	43.65&	42.37&	45.86\\
    CubeMLP \cite{10.1145/3503161.3548025}	&\multicolumn{1}{c|}{65.60}&	61.51&	60.82&	\multicolumn{1}{c|}{61.16}&	\multicolumn{1}{c|}{51.87}&	47.31&	45.07&	46.16\\ \hline
    \textbf{VC-CSA} &\multicolumn{1}{c|}{\textbf{73.52}}&	\textbf{67.51}&	\textbf{66.51}&	\multicolumn{1}{c|}{\textbf{69.19}}&	\multicolumn{1}{c|}{\textbf{62.99}}&	\textbf{55.18}&	\textbf{54.47}&	\textbf{56.36}\\  
    \hline
    \end{tabular} 
}
\label{tab_methods}
\end{table}
The performance metrics of each method is presented in Tab.~\ref{tab_methods} individually.
It is evident that the \textbf{VC-CSA} achieve the highest scores with a micro-F1 value of $73.52$ for opinion recognition and $62.99$ for emotion recognition.
It exhibits significant advantages over existing multi-modal methods in our proposed task, indicating the limitation of current approaches in addressing the distinctive challenges presented by our research.
Meanwhile, she results clearly demonstrate that multi-modal approaches outperform those depending solely on text, underscoring the importance of video content in interpreting sentiments of comments.



\textbf{Ablation Study.}
\begin{table}[t]
\centering
\caption{The Ablation study on our method. The Ablation Setting column is the alternative designs.}
\scalebox{0.8}{
\begin{tabular}{lllll}
\hline
Ablation Setting          & \begin{tabular}[c]{@{}l@{}}Opinion   \\   Micro F1\end{tabular} & \begin{tabular}[c]{@{}l@{}}Opinion   \\      Macro F1\end{tabular} & \begin{tabular}[c]{@{}l@{}}Emotion   \\      Micro F1\end{tabular} & \begin{tabular}[c]{@{}l@{}}Emotion   \\      Macro F1\end{tabular} \\ \hline
-Only single layer       & 72.35     & 65.51      & 62.06       & 54.18        \\
-Only last layer     & 69.13      & 63.37        & 59.67     & 51.81       \\
\begin{tabular}[c]{@{}l@{}}-LT  \end{tabular} & 72.32       & 66.43    & 62.52      & 54.74    \\
\begin{tabular}[c]{@{}l@{}}-AttnS  \end{tabular}       & 71.93       & 65.23   & 61.22    & 52.82     \\ \hline
\textbf{VC-CSA}      & \textbf{73.52}       & \textbf{67.51}   & \textbf{62.99}    & \textbf{55.18}   \\ \hline
\end{tabular}
}
\label{ablation}
\end{table}
We execute ablation studies on the three principal modules to validate the effectiveness. 
We adopted standard strategy instead of our custom design to assess performance difference. 
For the Multi-scale Temporal Representation, we use the \textbf{only single layer} and the \textbf{only last layer} CNN representation as the video feature at a single scale instead of it, respectively.
For the Consensus Semantic Learning, we replace our consensus token with the \textbf{last token in original transformer} (\textbf{LT} for short) encoder block. 
This involved concatenating the video and comment text and using the feature from the last position as the attention query.
For Golden Feature Grounding, we directly use the \textbf{attention score}  $S_{conAtt}$ (\textbf{AttnS} for short) as the grounding weight to obtain the video visual feature in relation to the comment.
The findings from the ablation studies are presented in Tab.~\ref{ablation}.
It is evident that excluding these designs from \textbf{VC-CSA} results in a decrease in evaluation metrics, highlighting their critical contribution to the method. 
A comparison between the ablated models and our complete model reveals an approximate $1-2\%$ improvement in the Micro F1 score.
More experiments discussions are available in the supplementary materials.

\section{Conclusion} \label{conclusion}
In conclusion, this study introduces the task of multi-modal sentiment analysis in comment of video-induced  (MSA-CRVI), focusing on understanding sentiment from comments related to micro-video content.
To support in this task, we have developed CSMV dataset, consisting of micro videos and their annotated comments.
The proposed VC-CSA method effectively infers sentiments from comments within the context of corresponding video, making a significant contribution for the novel multi-modal sentiment analysis setting.
Looking forward, we aim to enlarge the dataset and release more feature representations including audio features to further refine sentiment analysis capabilities.







\bibliographystyle{plainnat}
\bibliography{main-manu.bbl}

\appendix

\section{Dataset Details}
\label{sec:rationale}

\begin{figure}[]
    \centering
    \includegraphics[width=1\textwidth]{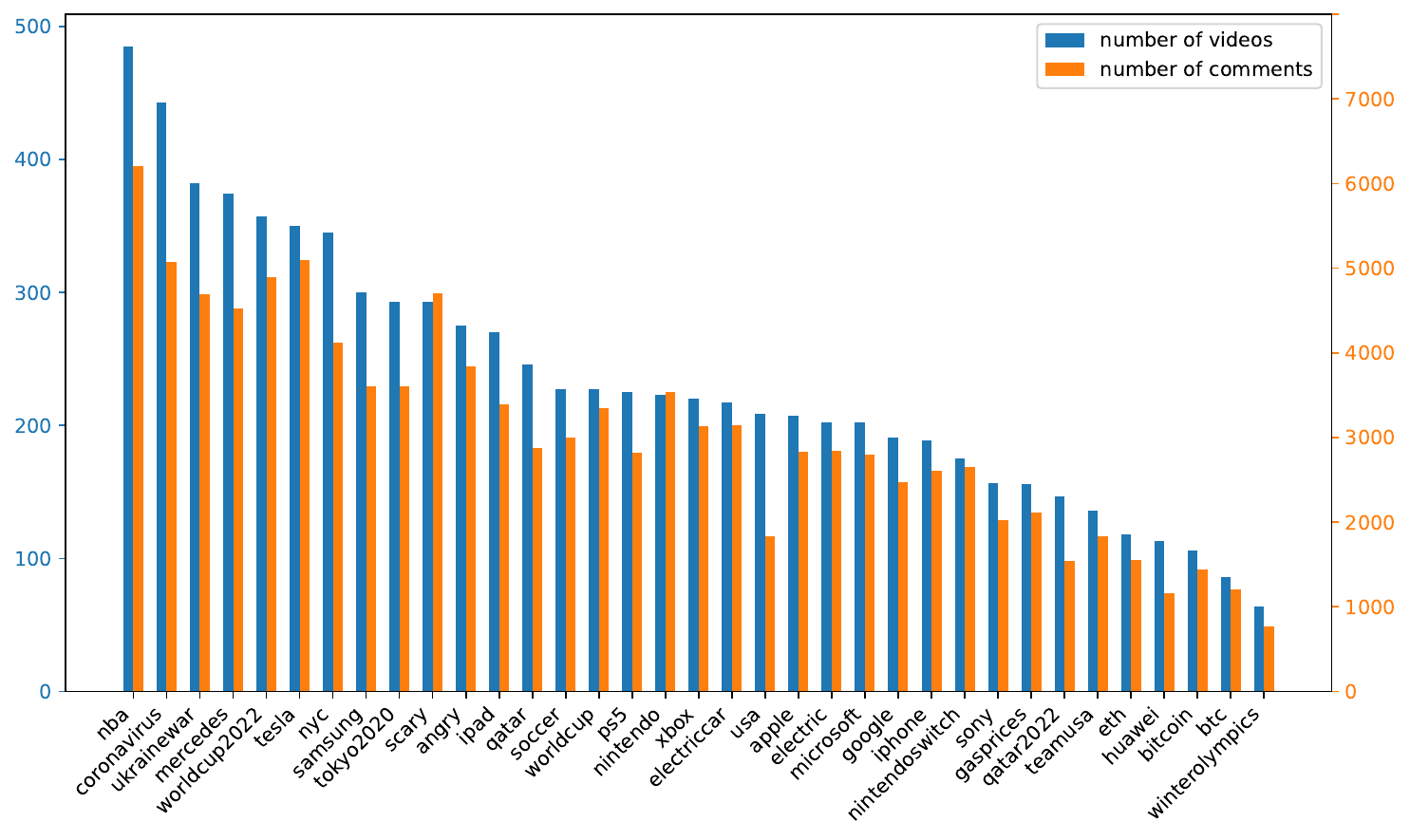}
    \caption{The distribution of the amounts of the micro video and comments under the hashtag.}
    \label{hash-video-comment}
\end{figure}

This section provides a comprehensive overview of the \textbf{CSMV} dataset.
The CSMV dataset comprises micro videos and their corresponding comments, which have been updated from February 2020 to October 2022.
This extensive time range allows for the inclusion of a diverse set of content, capturing the evolution of sentiments over the course of more than two years.
In total, the CSMV dataset comprises 8,210 micro videos, totaling approximately 68.83 hours of video duration, along with 107,267 related comments.
The CSMV dataset defines two distinct types of labels, opinion and emotion, for analyzing the sentiment expressed in the comments towards the micro videos.
By leveraging the combination of visual and textual content in this dataset, researchers are able to examine the interplay between language expressions and visual cues in determining sentiment.
To deepen our understanding of the CSMV dataset, we conducted an analysis of the distribution of videos and related comments using specific hashtags.
As depicted in Fig.~\ref{hash-video-comment}, this distribution exhibits a variety of topics, suggesting a rich diversity of sentiments and experiences captured within the dataset.
Moreover, this diversity enhances the dataset’s versatility and utility for multi-modal sentiment analysis tasks.
By leveraging the CSMV dataset, researchers and practitioners can delve into multi-modal sentiment analysis, exploring the intricate relationships between visual content and textual expressions. 
This comprehensive understanding not only aids in opinion classification and emotion detection but also provides valuable insights into the complex interplay of modalities within human communication.

\begin{figure}[]
    \centering
    \subfloat[Opinion label distribution.]{\includegraphics[width=0.5\linewidth]{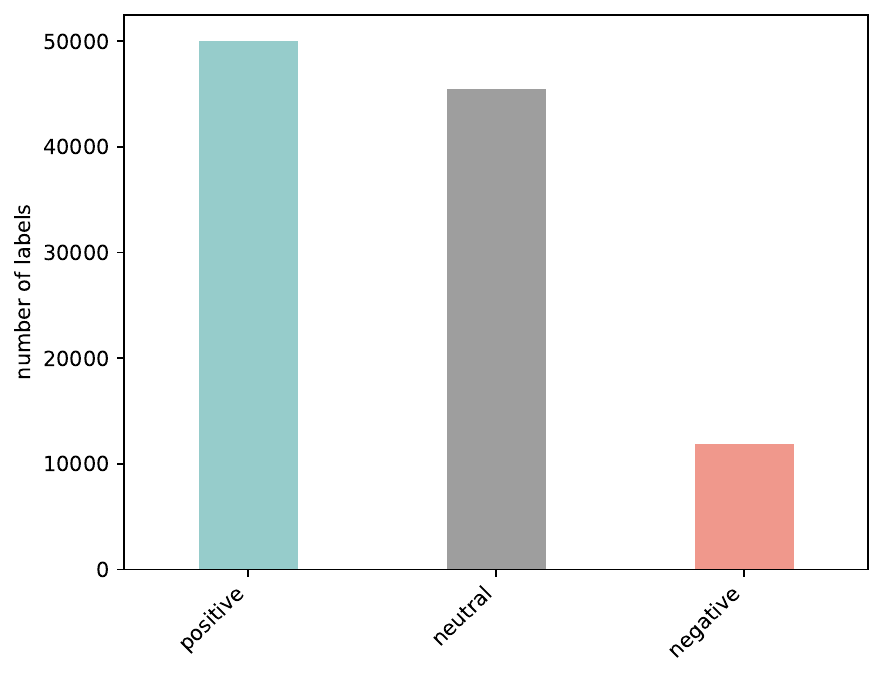}\label{old}}
    \hfill
    \subfloat[Emotion label distribution.]{\includegraphics[width=0.5\linewidth]{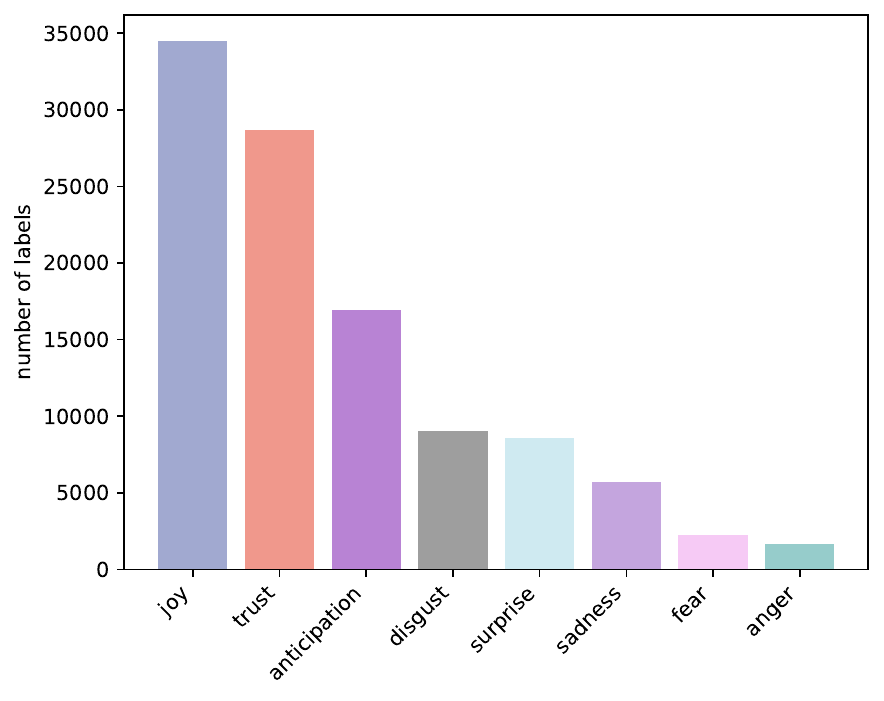}\label{eld}}
    \caption{The distribution of the number of labels in our CSMV dataset.}
    \label{dataset-1}
\end{figure}

The distribution of labels in our CSMV dataset is shown in Fig.~\ref{dataset-1}.
In Fig.~\ref{old}, the opinion labels are distributed as follows: positive - 47\%, neutral - 42\%, and negative - 11\%.
Negative comments are clearly in the minority.
The distribution of emotion labels is depicted in Fig.~\ref{eld}.
According to the statistical results, the top two labels with the highest proportion are joy (32\%) and trust (27\%).
Additionally, the labels of sadness, fear, and anger have the smallest proportions, accounting for 5\%, 2\%, and 2\%, respectively.
The distribution indicates an imbalance in labels.
The distribution of opinions aligns with users’ behavioral tendencies in the comments on micro videos.
Users are more inclined to write comments providing positive feedback.
Conversely, expressing negative opinions about videos they dislike is considered rare, and users tend to ignore such micro videos.
TikTok uploaders are more inclined to create and share content to attract other users.
Consequently, the majority of comments have a non-negative sentiment.
The distribution of emotion labels follows a similar pattern.
On the social media platform, users tend to express positive emotions, such as joy and trust, more frequently than negative ones.
Based on these findings, our dataset accurately represents the real distribution of human sentiment in real-world scenarios.

\begin{figure}[]
    \centering
    \includegraphics[width=0.75\linewidth]{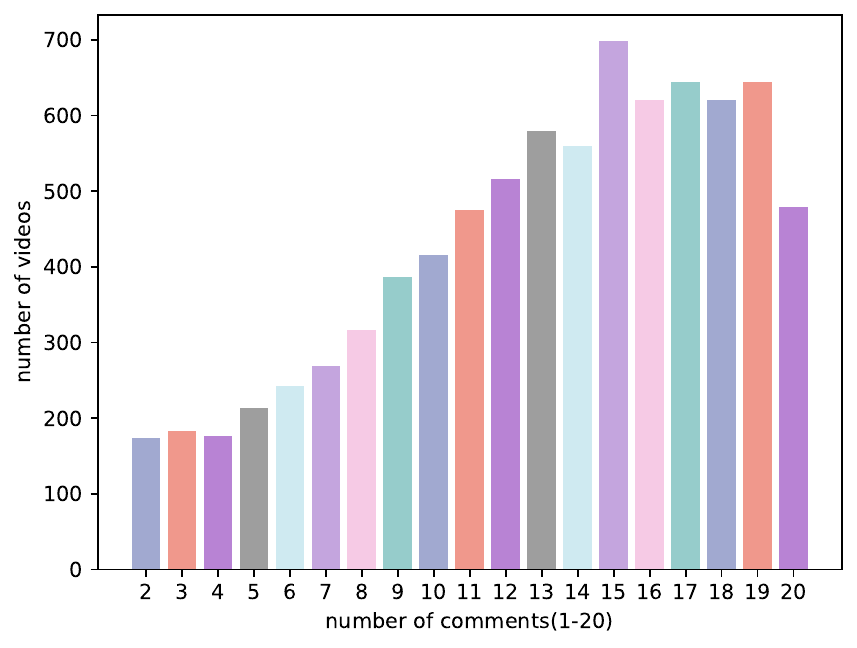}
    \caption{The distribution of the amounts of the comments under single micro video.}
    \label{dvcg}
\end{figure}

One primary objective of the CSMV dataset is to enhance the diversity of sentiment responses towards videos. 
This goal is achieved through the annotation of multiple comments for each video, ensuring a broader range of sentiment expressions within CSMV.
An analysis of our dataset has been conducted, and the results are discussed in Fig.~\ref{dvcg}. 
The distribution clearly illustrates that the majority of our videos provide more than 10 annotated comments, while only a small proportion have 2-5 annotated comments.
This observation signifies that our dataset exhibits a greater level of complexity compared to conventional multi-modal sentiment analysis tasks. 
By incorporating multiple comments for each video in our annotation process, we enable a more comprehensive and nuanced understanding of human sentiment. 
This approach facilitates the training of artificial intelligence systems to recognize and respond to diverse human experiences, as our dataset CSMV encompasses a broader spectrum of sentiment responses. 
The inclusion of multiple comments allows for a deeper exploration of the various opinions and emotions conveyed by individuals in response to videos. 
Each comment represents a distinct opinion, contributing to a rich complexity of sentiment found within CSMV. 
In essence, the CSMV dataset's commitment to incorporating a diverse range of sentiment responses strengthens its capacity to train AI systems, which make it capable of accurately identifying and effectively responding to the multitude of sentiment expressed by individuals in relation to videos.
This comprehensive collection of sentiment responses enables researchers and developers to develop a more sentimental AI system.
It could possess a more profound understanding of human sentiment and own the ability to discern and appropriately react to the intricacies of human experience.

\begin{figure}[]
    \centering
    \subfloat[Video length distribution.]{\includegraphics[width=0.5\linewidth]{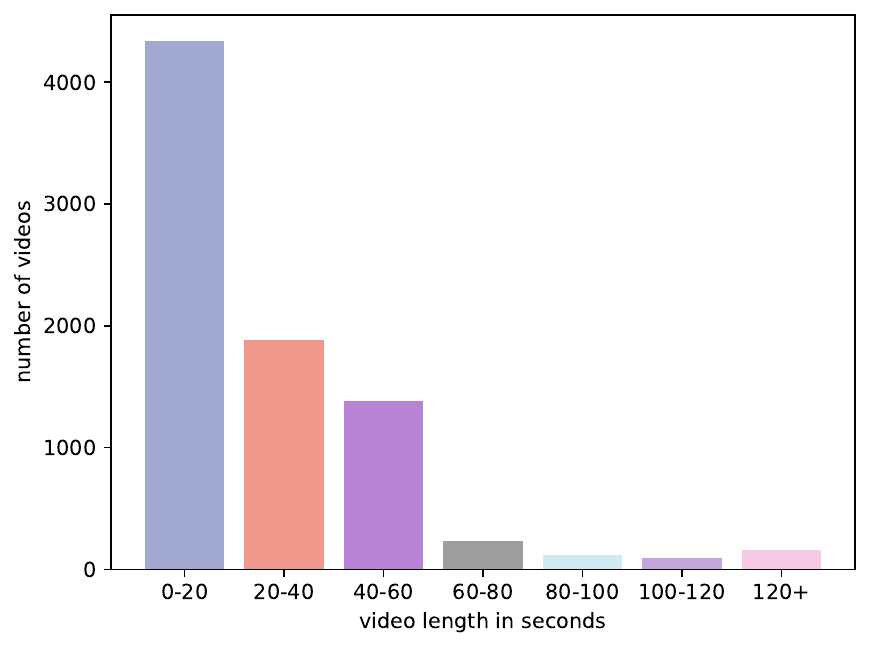}}
    \hfill
    \subfloat[Comment length distribution.]{\includegraphics[width=0.5\linewidth]{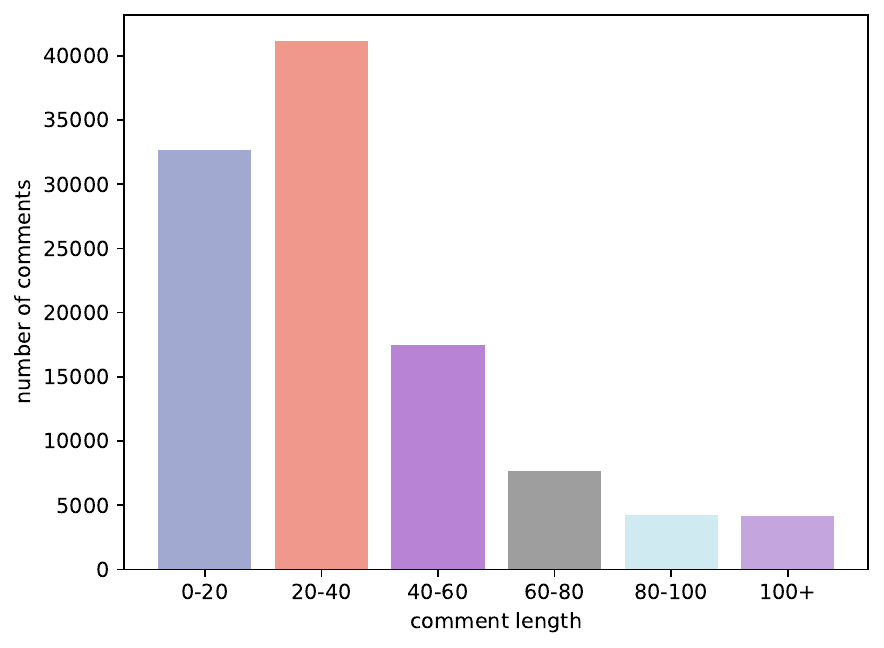}}
    \caption{The distribution of the length of video duration and comment text.}
    \label{VL_CL-d}
\end{figure}

Length information for the samples in the CSMV dataset, is mainly focus on the distribution of video duration and comment text length.
The analysis, as depicted in Fig.~\ref{VL_CL-d} (a), indicates that the majority of videos have a duration of 60 seconds or less, with the highest proportion falling between 0 and 20 seconds.
Micro video creators aim to convey impactful narratives within a limited time frame, enabling the popularity of their content on social media platforms.
Likewise, comments in our dataset follow a distribution pattern similar to that of video duration.
Fig.~\ref{VL_CL-d} (b) illustrates that the majority of comments contain 60 characters or less, with the most frequently observed length ranging from 40 to 60 characters.
When users respond to a video by commenting, they may use concise text with video-specific unconventional abbreviations.
Without the visual context provided by the video, understanding the intended message can be difficult. 
In conclusion, analyzing the video duration and comment length in our CSMV dataset emphasizes the prevalence of concise communication and short-form content in the field of social media.
Utilizing brevity and unconventional abbreviations, creators and commenters endeavor to captivate audiences and engage in fast-paced online discourse.
A comprehensive understanding of the underlying context and meaning of these micro videos and comments necessitates video context support.

\section{Optimization Hyper-parameter Experiment}

Our \textbf{VC-CSA} integrates multiple modules requiring hyper-parameters configuration. 
To determine the  setup, we explore the several key factors through experiments: 
(1) the layer count in Multi-scale Temporal Representation module;
(2) the layer count in Consensus Transformer module corresponding to each scale of multi-scale temporal representations $\{{f_v^i}\} $, and 
(3) the count of Consensus Tokens in the Consensus Transformer. 
Systematic experiments  with these parameters as shown in Tab.~\ref{hyperP}. 
The findings indicate that the VC-CSA model achieves optimal performance with 4 layer Multi-scale Temporal representation, 1 layer Consensus Transformer, and 1 Consensus Token.
\begin{table}[ht]
\centering
\scalebox{0.8}{
\begin{tabular}{lllll}
\cline{1-5}
\begin{tabular}[c]{@{}l@{}}Hyper-parameter \\ setting \end{tabular}                                                                  & \begin{tabular}[c]{@{}l@{}}Opinion \\ Micro F1\end{tabular} & \begin{tabular}[c]{@{}l@{}}Opinion \\ Macro F1 \end{tabular}& \begin{tabular}[c]{@{}l@{}}Emotion \\ Micro F1 \end{tabular}& \begin{tabular}[c]{@{}l@{}}Emotion \\ Macro F1 \end{tabular} \\ \cline{1-5}
2,1,1 & 72.62                                                       & 66.49                                                       & 62.10                                                       & 53.85                                                       \\
3,1,1 & 71.79                                                       & 65.78                                                       & 60.82                                                       & 52.58                                                       \\
\textbf{4,1,1} & \textbf{73.52}                      & \textbf{67.51}                      & \textbf{62.99}                      & \textbf{55.18}                      \\ 
5,1,1 & 72.67                                                       & 66.38                                                       & 62.69                                                       & 54.90                                                       \\
6,1,1& 72.46                                                       & 66.32                                                       & 61.93                                                       & 54.26                                                       \\ \hline
4,1,2 & 70.83                                                       & 64.11                                                       & 60.88                                                       & 52.41                                                       \\
4,1,3 & 69.42                                                       & 63.62                                                       & 59.80                                                       & 50.66                                                       \\\hline
4,2,1 & 72.58                                                       & 65.99                                                       & 62.52                                                       & 54.64                                                       \\
4,3,1 & 72.72                                                       & 65.81                                                       & 61.99                                                       & 53.27                                                       \\  \cline{1-5}
\end{tabular}
}
\caption{The experimental results of diverse hyper-parameter settings for \textbf{VC-CSA}. In the Hyper-parameter setting column, the three parameters are the layer count in Multi-scale Temporal Representation module, the layer count in the Consensus Transformer module, and the count of Consensus Tokens.}
\label{hyperP}
\end{table}

\section{The Importance of Video in MSA-CRVI Task}
\begin{figure}[]
  \centering
  \includegraphics[width=1\linewidth]{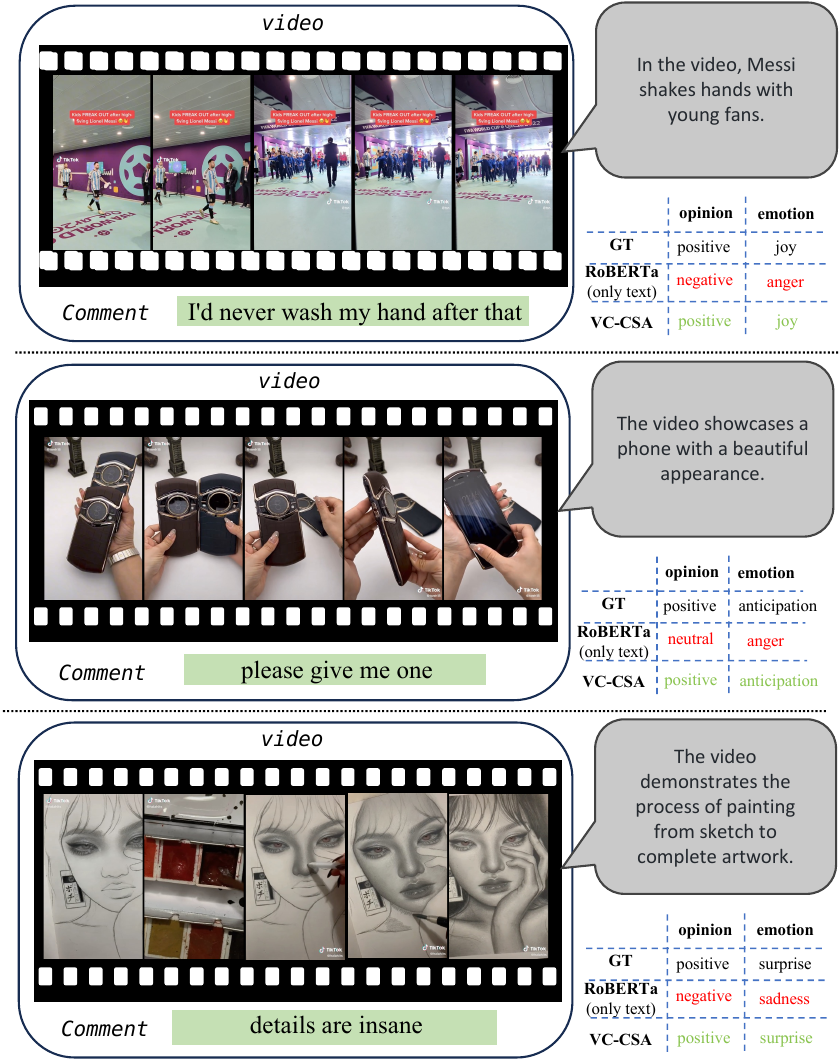} 
  \caption{The figure shown the inference results of several test set samples, including the ground truth, inference results of RoBERTa (only text), and our VC-CSA method.} 
  \label{samples} 
\end{figure}

To demonstrate the significance of video in the MSA-CRVI task, we present the inference outcomes for several samples from the test set, as shown in Fig.~\ref{samples}.
A shared feature among the three samples is that the absence of video information complicates the interpretation of the intended meaning when relying solely on textual comments. 
This often leads to inaccurate inferences about the opinions and emotions conveyed in the comments related to the video.
Even the fine-tuned  RoBERTa model on the comment of the CSMV dataset exhibits notable deviations as well.
However, our VC-CSA model demonstrates this capability by accurately identifying sentiment within the context of video content, proving the integration of video viewing enables a more accurate understanding of the comments.

\begin{figure}[]
    \centering
    \begin{minipage}[b]{1\linewidth}
    \subfloat[Example 1]{
        \begin{minipage}[b]{0.5\linewidth}
            \centering
            \includegraphics[width=\linewidth]{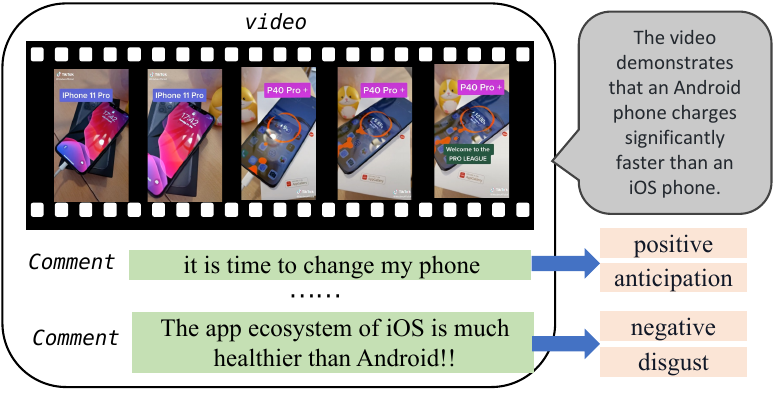}
        \end{minipage}
        \label{examples1}
    }
    \hfill
    \subfloat[Example 2]{
        \begin{minipage}[b]{0.5\linewidth}
            \centering
            \includegraphics[width=\linewidth]{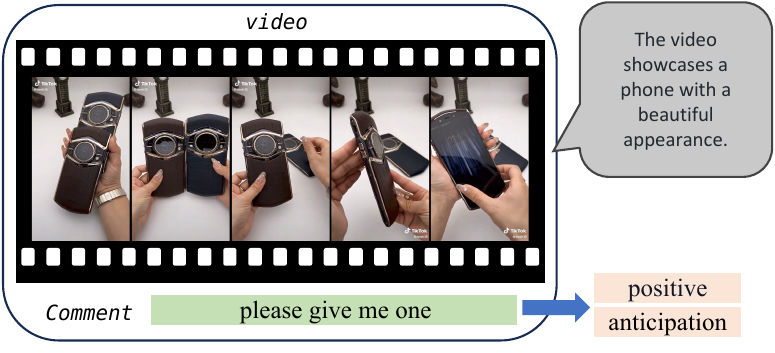}
        \end{minipage}
        \label{examples2}
    }
    \hfill
    \subfloat[Example 3]{
        \begin{minipage}[b]{0.5\linewidth}
            \centering
            \includegraphics[width=\linewidth]{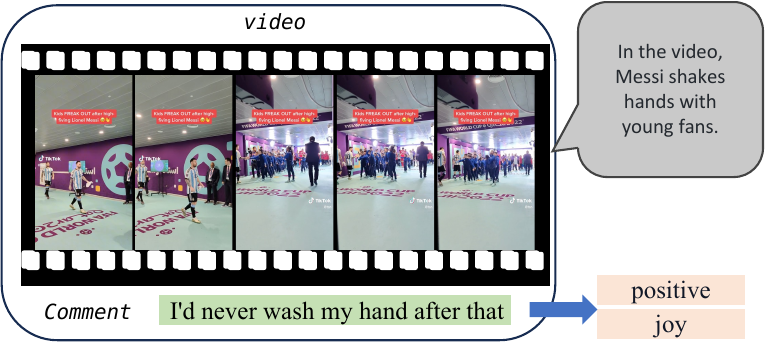}
        \end{minipage}
        \label{examples3}
    }
    \subfloat[Example 4]{
        \begin{minipage}[b]{0.5\linewidth}
            \centering
            \includegraphics[width=\linewidth]{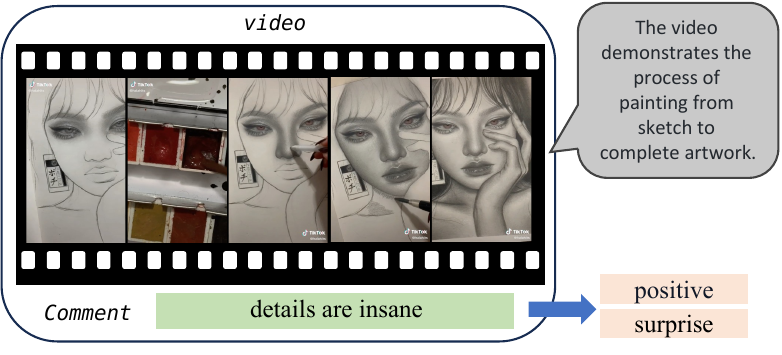}
        \end{minipage}
        \label{examples4}
    }
    \end{minipage}
    \vfill
\caption{Some examples in the CSMV dataset. To enhance comprehension, a concise description of the video content is presented in a gray box. This description does not serve as input.
}
\label{4examples}
\end{figure}


\textbf{Samples Explanation.} To provide a comprehensive explanation of our task, we will offer detailed explanations for the four selected examples discussed in the main article, as shown in Fig.~\ref{4examples}.

Example 1 in Fig.~\ref{examples1} highlights a common debate in the smartphone world-Android versus iOS.
The video showcases a comparison between the charging speed of an Android smartphone and an iOS smartphone, aiming to demonstrate the superiority of Android over iOS. 
The comment “it is time to change my phone.” in response to the video expresses agreement with its content and a desire to acquire a new phone.
However, another comment argues that advocating Android’s superiority based solely on charging speed is a one-sided argument and points out that the iOS platform has a healthier application ecosystem, stating “The app ecosystem of iOS is much healthier than Android!”
This comment conveys a negative opinion towards the video content and a sense of disgust.
This contrasting opinion reveals the diverse perspectives within the audience.

Moving on to Example 2 in Fig.~\ref{examples2}, the video focuses on smartphones and presents a visually appealing device. 
A comment, “please give me one,” acknowledges the aesthetic appeal of the phone shown in the video and expresses a desire to possess it.
This comment not only highlights the allure of visually appealing products but also emphasizes the psychological impact of such imagery on consumer behavior. 
It demonstrates how videos can act as persuasive tools in influencing audiences’ desires.

In Fig.~\ref{examples3}, the video depicts Lionel Messi walking through the player tunnel and shaking hands with young fans on the opposite side.
An interesting comment in response states, “I’d never wash my hand after that.” This comment expresses the idea that shaking hands with Messi is such a joyful experience that they would not want to wash their hand afterward, in order to preserve that moment.
It represents a positive reaction to the video content and conveys joy.
This reaction illustrates how videos with positive or emotionally charged content can evoke strong emotional responses from audience, emphasizing their power to create a sense of connection and emotional engagement.

Lastly, Example 4 presented in Fig.~\ref{examples4} showcases a video of a painting process starting with a rough sketch and gradually completing the entire artwork.
A comment exclaims, “details are insane,” expressing astonishment at the level of detail in the painting process and acknowledging the outstanding intricacy of the artwork.
This comment reflects a positive reaction to the video and acknowledges the presented impressive level of detail.
This response not only offers an endorsement of the video’s content but also serves as an acknowledgement of the skill and talent of the artist.

Through the aforementioned examples, we can reiterate the objective of our study.
Our task entails inferring the opinions toward to the content of the video, as well as the emotions evoked by it.
We need to consider the video as the contextual backdrop and pay attention to the intricate interaction between the comments and the video in order to ensure precise inference.
These examples shed light on the multifaceted nature of videos and their ability to influence opinions and evoke emotions.
Analyzing the comments alongside the videos helps us understand the diverse perspectives, desires, and reactions of the audiences.
Understanding the intricate interaction between videos and comments provides valuable insights into the impact and effectiveness of video content in online discourse.

\end{document}